\documentclass[conference]{IEEEtran}

\usepackage{cite}
\usepackage{amsmath,amssymb}
\usepackage{graphicx}
\usepackage{float}
\usepackage{algorithm2e}
\usepackage{booktabs}
\usepackage{multirow}
\usepackage{placeins}
\usepackage{wrapfig}
\usepackage[export]{adjustbox} 
\usepackage{tabularx,array}

\newcommand{\authorEmailSize}{\scriptsize}

\title{Optimal Transport for handwritten text recognition in a low-resource regime}

\author{
    \IEEEauthorblockN{
        Petros Georgoulas Wraight\textsuperscript{1,4}, 
        Giorgos Sfikas\textsuperscript{3}, 
        Ioannis Kordonis\textsuperscript{2}, 
        Petros Maragos\textsuperscript{1,2,4} and 
        George Retsinas\textsuperscript{1}
    }
    \IEEEauthorblockA{
        {\authorEmailSize\texttt{p.georgoulas@athenarc.gr, gsfikas@uniwa.gr, kordonis@central.ntua.gr, petros.maragos@athenarc.gr, gretsinas@athenarc.com}}\\
        \vspace{1em} 
        \begin{tabularx}{0.92\textwidth}{>{\centering\arraybackslash}X}
        \textsuperscript{1}Institute of Robotics, Athena Research Center, Maroussi, Greece\\
        \textsuperscript{2}School of ECE, National Technical University of Athens, Greece\\
        \textsuperscript{3}Department of SGE, University of West Attica, Athens, Greece\\
        \textsuperscript{4}HERON - Hellenic Robotics Center of Excellence, Athens, Greece
        \end{tabularx}
    }
}

\def\BibTeX{{\rm B\kern-.05em{\sc i\kern-.025em b}\kern-.08em
    T\kern-.1667em\lower.7ex\hbox{E}\kern-.125emX}}

\begin{document}
\maketitle

\begin{abstract}
Handwritten Text Recognition (HTR) is a task of central importance in the field of document image understanding.
State-of-the-art methods for HTR
require the use of extensive annotated sets for training, making them impractical for low-resource domains like historical archives or limited-size modern collections.
This paper introduces a novel framework that,
unlike the standard HTR model paradigm,
can leverage mild prior knowledge of lexical characteristics;
this is ideal for scenarios where labeled data are scarce.
We propose an iterative bootstrapping approach that aligns visual features extracted from unlabeled images with semantic word representations using Optimal Transport (OT).
Starting with a minimal set of labeled examples, the framework iteratively matches word images to text labels, generates pseudo-labels for high-confidence alignments, and retrains the recognizer on
the growing
dataset.
Numerical experiments demonstrate that our iterative visual-semantic alignment scheme significantly improves recognition accuracy on low-resource HTR benchmarks.
\end{abstract}



\section{Introduction}

Handwritten Text Recognition (HTR) plays a pivotal role in the digital humanities by enabling the automated transcription and analysis of vast collections of historical manuscripts, letters, and archival documents that would otherwise remain inaccessible or require labor-intensive manual effort. This technology facilitates the preservation and democratization of cultural heritage, allowing researchers to search, index, and interpret handwritten texts at scale. For instance, HTR has been instrumental in digitizing historical archives, transforming degraded or complex documents into machine-readable formats that support scholarly editions and computational analysis \cite{sanchez2021handwritten}.

Traditional approaches rely on extensive annotated datasets to train recognition models, necessitating meticulous manual labeling 
of text sequences within images, 
a process that is not only labor-intensive but also expensive.
This is particularly the case
for large-scale datasets where 
experts must transcribe degraded or archaic text, often leading to inconsistencies and errors in ground-truth (GT) data \cite{sanchez2021handwritten}. This 
on supervised learning 
scheme 
is especially impractical for low-resource domains, such as historical archives, where annotated data are scarce, costly to produce, or simply unavailable due to the specialized knowledge required for accurate transcription.

In this work, we aim to overcome these limitations by casting HTR 
as visual–semantic matching and propose an iterative bootstrapping framework that aligns visual descriptors of unlabeled word images with semantic word representations. A lexical prior is assumed, in the form of knowledge about valid word instances and relative frequencies in the target vocabulary. This provides the distribution over the semantic space, onto which images from the visual space are to be aligned.
Starting from a small set of seed word–image pairs, 
a backbone produces visual descriptors that a small MLP projector maps into the word-embedding space; 
each round solves an Optimal Transport (OT) problem between the projected descriptors and the vocabulary embeddings to obtain a distribution-level coupling that minimizes transport cost. 
The most confident matches are pseudo-labelled and the recognizer is retrained on the expanded set, so alignment and recognition improve progressively. This key novelty,
namely reframing HTR as OT-driven visual-semantic alignment with iterative pseudo-label expansion, enables robust automatic transcription with minimal training GT requirements.

In a nutshell, with this paper our contribution is two-fold:
\begin{itemize}
    \item We introduce a novel model paradigm for HTR,
    that casts the problem under an intuitive Optimal Transport-based self-training scheme;
    our results show that our model
    is especially ideal in a resource-constrained setting.
    \item We show that, unlike standard HTR models, our model can leverage lexical prior knowledge to obtain a considerable boost in performance (up to more than $10\%$ of improvement over the current state of the art).
\end{itemize}

\section{Related Work}
\label{sec:related}
Optimal Transport (OT) provides a principled mechanism to align probability measures across tasks. The Earth Mover’s Distance established its perceptual relevance for image similarity \cite{rubner2000earth}; OT couplings have since supported domain adaptation by matching source and target feature distributions in both classical and deep end-to-end formulations 
\cite{ott2022domain}. 
OT also propagates sparse supervision and induces correspondences in segmentation and few-shot settings 
\cite{liu2022few}, 
and improves label assignment in detection \cite{de2023unbalanced,ge2021ota}. Beyond supervision transfer, partial OT has been used for label mapping in visual prompting \cite{zheng2024visual}.

In document analysis and HTR, OT has been employed mainly to shift feature distributions across writers or domains to mitigate mismatch \cite{ott2022domain}, and as a string-similarity mechanism aligning learned character representations \cite{tam2019optimal}. However, these uses do not tackle low-label transcription, 
via explicit cross-modal alignment between word-image descriptors and lexical representations. 
Our formulation addresses this, 
by making OT coupling itself the generator of pseudo-labels.

Low-resource HTR contends with scarce annotations, pronounced style/domain shift, and script-specific hurdles. Some of the effective strategies combine synthetic pretraining with transductive adaptation to target collections \cite{keret2019transductive}, and decoder-side transfer mapping letter $n$-grams to words across languages and periods \cite{granet2018transfer}. Most relevant are self-training pipelines that iteratively annotate unlabeled word images and retrain, yielding steady gains \cite{wolf2022combining, wolf2022self}. Our approach follows this paradigm but differs in how pseudo-labels are produced and vetted: we cast recognition as visual-semantic alignment and use OT to transport projected image descriptors onto a word-embedding manifold before promotion. 








\begin{figure}
    \centering
    \includegraphics[width=0.99\linewidth]
    {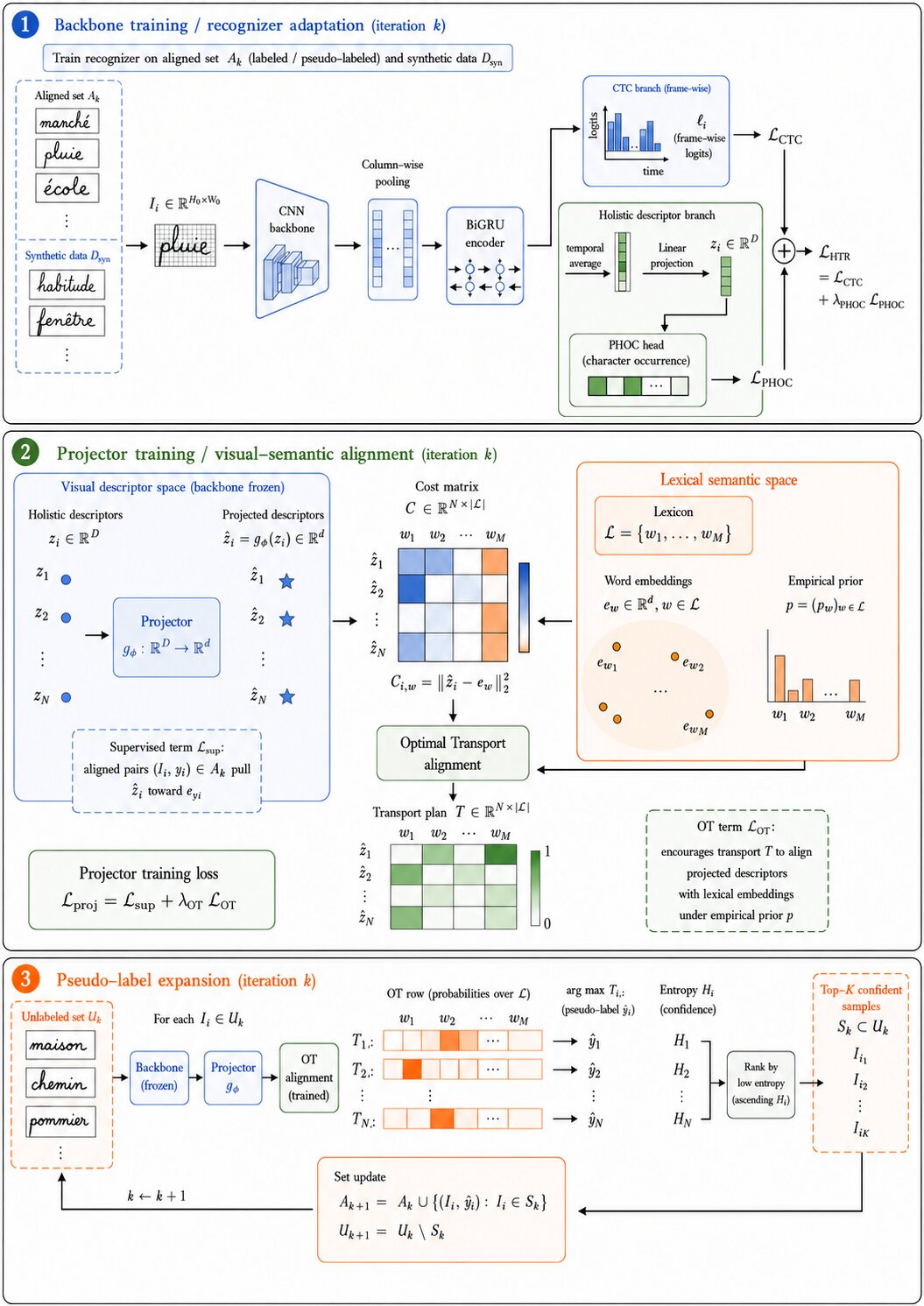}
    \vspace{-5pt}
    \caption{Overview of one iteration of the proposed visual--semantic alignment loop. 
Phase A adapts the CNN--BiGRU recognizer on the aligned set $A_k$ using CTC and PHOC losses. 
Phase B freezes the backbone and trains a projector $g_\phi$ that aligns holistic image descriptors with lexical embeddings through entropic Optimal Transport under the empirical prior $p$. 
Phase C converts low-entropy OT assignments into top-$K$ pseudo-labels, expanding $A_k$ and shrinking $U_k$ for the next iteration.}
    \label{fig:placeholder}
\end{figure}

\section{Proposed Model and Method} \label{sec: proposed method}

\textbf{Overview.}
\label{subsubsec: overview}
With the proposed framework, we exploit a basic regularity of language:
certain 
words (e.g., “and”, “of”) occur far more often than 
others
(e.g., “philosopher”), 
a Zipfian skew that should manifest as denser regions in a well‑behaved visual feature space for handwritten words \cite{vidal2024zipf}. We cast HTR as cross‑modal matching between a \emph{visual space} of image descriptors and a 
\emph{word embedding space} of lexical tokens.
The key notion to this end is Optimal Transport (OT), 
with which we align their distributions under minimal supervision,
enabling robust recognition with lexicon-guided training and lexicon-free inference. 


To instantiate this matching, we maintain two complementary spaces—a visual space for word‑image descriptors and a lexical embedding space for candidate tokens—together 
with a pretraining-initialized mapping
between them. 
On the visual side, a compact recognizer produces (i) frame‑wise logits trained with the Connectionist Temporal Classification (CTC) loss, which we use for lexicon‑free decoding at inference, and (ii) a single global descriptor formed by aggregating the sequence states over time, placing each image as a point in the visual space. A PHOC 
\cite{architecture_retsinas}
auxiliary prediction head
acts as a mild regularizer, 
encouraging descriptors that share character content to lie near one another, since PHOC is a pyramidal representation of character occurrence. 
These representations provide the geometric substrate that the subsequent projectors and OT alignment operate on.

The word embedding space
is built by manifold learning over a candidate vocabulary: we compute pairwise Levenshtein distances 
among lexicon entries and apply Multi‑Dimensional Scaling (MDS) \cite{mds} to embed them into a low‑dimensional Euclidean space that preserves lexical proximity.

End‑to‑End, each image yields both frame‑level logits and a global visual descriptor.
The latter is mapped into the 
word embedding space
by a lightweight \emph{projector component}. OT then aligns the projected visual descriptors with word embeddings and yields transport plans used for pseudo-label assignment, while the CTC decoder produces the final, lexicon-free transcriptions. High-confidence matches expand the aligned set and are used to retrain the recognizer, progressively refining both the backbone and subsequent alignment steps. OT is used only to guide pseudo-label selection during training; at inference, we omit OT and decode directly from the CTC logits, keeping the method lexicon-free. Using OT at test time would require fixing a candidate lexicon \emph{a priori} (to define the embedding space and target marginal), which is often unavailable in real-world applications.

\noindent\textbf{Proposed Architecture.} Our network follows the compact residual CNN of Retsinas et al. \cite{architecture_retsinas} with adaptations for the iterative alignment framework. An input image $I\in\mathbb{R}^{H_0\times W_0}$ is mapped to a feature map $Y\in\mathbb{R}^{C\times H\times W}$ through convolutional stacks with max-pooling. As in \cite{architecture_retsinas}, a column-wise global max-pool with kernel $(H,1)$ collapses vertical variation and yields a sequence of $W$ column descriptors $\{f_w\}_{w=1}^W,f_w\in\mathbb{R}^C$. This sequence feeds two branches. 

\textit{Sequential transcription head.} The descriptors $S = [f_1,\dots,f_W]\in\mathbb{R}^{W\times C}$ are processed by a bidirectional GRU (hidden size $H_{\text{GRU}}$) to produce states $h_w = [h_{\overrightarrow{w}} ; h_{\overleftarrow{w}}]\in\mathbb{R}^{2H_{\text{GRU}}}$. A linear layer maps each $h_w$ to logits $\ell_w\in\mathbb{R}^{|\Sigma|}$ over the alphabet $\Sigma$. 

\textit{Holistic descriptor.} The same recurrent states are averaged and linearly projected to form a global image vector $z = B\left(\frac{1}{W}\sum_{w=1}^W h_w\right)\in\mathbb{R}^D$, where $B\in\mathbb{R}^{D\times 2H_{\text{GRU}}}$. This descriptor serves the subsequent cross‑modal alignment.

\textit{PHOC auxiliary head.} To inject lightweight linguistic structure, a single fully connected layer predicts a PHOC vector at multiple pyramid levels from $z$ 
\cite{architecture_retsinas}.
This auxiliary prediction encourages $z$ to encode character‑presence patterns, promoting semantically coherent neighborhoods that benefit the later alignment stage.

\noindent\textbf{Stage I – Synthetic Pre‑Training.} We apply a brief warm-start on a small subset of the \textsc{Synth90k} corpus \cite{MJSynth}: the sequential transcription head learns a basic mapping and the global image descriptor is organized enough that the first pseudo‑labelling step has non‑random geometry to work with. No projector training is involved in this stage. We use the same recognition objective introduced in Phase A (Eq. \ref{eq: backbone_loss}).

\label{subsec:pretraining}

\noindent\textbf{Stage II - Iterative Visual–Semantic Alignment.}
At iteration $k$, the training pool is split into aligned items $\mathcal{A}_k\subset\mathcal{I}_{\text{train}}$ (with known/estimated labels) and unaligned items $\mathcal{U}_k = \mathcal{I}_{\text{train}}\setminus\mathcal{A}_k$ (unknown labels). The goal of each round is to enlarge $\mathcal{A}_k$ and shrink $\mathcal{U}_k$ before proceeding to $k+1$. The loop has three phases, summarized below. 

\emph{Phase A - Backbone Adaptation:}
\label{subsubsec:phaseA}
We fine‑tune the backbone by sampling mini-batches that consist of a mixture of synthetic and real aligned samples. Specifically, each batch is composed with proportion $\rho_{\text{syn}}$ drawn from the synthetic dataset $\mathcal{D}_{\text{syn}}$ and proportion 
$1 - \rho_{\text{syn}}$
drawn for the aligned set $\mathcal{A}_k$, optimizing the composite loss 
\begin{equation} \label{eq: backbone_loss}
\mathcal{L}_{\text{HTR}} = \mathcal{L}_{\text{CTC}} + \lambda_{\text{PHOC}}\mathcal{L}_{\text{PHOC}},
\end{equation}
where \(\mathcal{L}_{\mathrm{CTC}}\) enforces character-level consistency along the temporal sequence while \(\mathcal{L}_{\mathrm{PHOC}}\) drives the global descriptor $z$ to predict the multiscale character histogram of its own transcription.  The first term teaches the network how letters assemble into words; the second injects explicit character-presence signals directly into the embedding.  Together they sculpt an initial visual manifold in which distances already reflect lexical relationships, providing a stable foundation for the iterative alignment stages that follow.

\emph{Phase B - Projector Training:} 
\label{subsubsec: phaseB}
At iteration $k$, we freeze the visual backbone and train a lightweight projector $g:\mathbb{R}^D\rightarrow\mathbb{R}^d$ that maps each image descriptor $z_i\in\mathbb{R}^D$ into the 
word
embedding space spanned by $\{e_w\in\mathbb{R}^d:w\in\mathcal{L}\}$,
where $\mathcal{L}$ denotes our lexicon and $p = (p_w)_{w\in\mathcal{L}}$
its empirical frequency prior.
Given $Z_k = \{z_i\}_{i=1}^N$ as the set of holistic descriptors computed for all images under consideration at iteration $k$, 
we write $\hat{z}_i = g(z_i)$ and define the empirical measure $\mu_k = \frac{1}{N}\sum_{i=1}^N\delta_{\hat{z}_i}$,
where $\delta_{j}$ is a Dirac point mass element centered on $j$.
The lexical measure is $\nu = \sum_{w\in\mathcal{L}}p_w\delta_{e_w}$. 

\begin{enumerate}
    \item \textbf{Supervision on labeled pairs.} For aligned items $(I_i,y_i)\in\mathcal{A}_k$, we penalize the deviation between the projected descriptor and the embedding of its transcription:
    \begin{equation}
        \mathcal{L}_{\text{sup}} = \frac{1}{|\mathcal{A}_k|}\sum_{(I_i,y_i)\in\mathcal{A}_k} ||\hat{z}_i - e_{y_i}||_2^2
    \end{equation}

    \item \textbf{Entropic OT to the Lexicon.} To globally align $\mu_k$ with $\nu$, we define the feasible set of couplings
        \begin{equation}\label{eq: feasible_ot_set}
\begin{split}
\!\!\!\!\!\!\!\!
\Pi = \{T\in\mathbb{R}^{N\times|\mathcal{L}|}: T\geq 0,\;
T\mathbf{1}_{|\mathcal{L}|}=\tfrac{1}{N}\mathbf{1}_N,
T^T\mathbf{1}_N=p\}
\end{split}
\end{equation}
    the entropically-regularized OT objective
    \cite{de2023unbalanced,ott2022domain}
    is then 
    \begin{equation}
        \mathcal{L}_{\text{OT}} = \min_{T\in\Pi}\langle T,C\rangle + \epsilon\sum_{i,w} T_{i,w}(\log T_{i,w} - 1)
            \vspace{-0.2cm}
    \end{equation}
    where the cost matrix $C\in\mathbb{R}^{N\times|\mathcal{L}|}$ contains squared Euclidean distances $C_{i,w} = ||\hat{z}_i - e_w||_2^2$. Here $\epsilon$ controls the strength of the regularizer and $\langle\cdot,\cdot\rangle$ denotes the Frobenius inner product. 
\end{enumerate}

The projector $g$ is trained with $\mathcal{L}_{\text{proj}} = \mathcal{L}_{\text{sup}} + \lambda_{\text{OT}}\mathcal{L}_{\text{OT}}$.

\emph{Phase C - Pseudo‑Label Expansion:} \label{subsubsec:phaseC}
Phase C turns the soft alignments from Phase B into new training labels. The OT solver provides, for each unlabeled image, a distribution over all candidate words indicating how strongly the image matches each entry in the lexicon. We interpret this distribution as confidence scores: concentrated mass on a single word signals a confident match, whereas a spread-out distribution indicates uncertainty. To quantify this, we measure the uncertainty of each image using the Shannon entropy of its distribution; lower entropy means higher confidence.

At each iteration, we sort the unlabeled pool from most to least confident and select a fixed budget of top $K$ candidates. We use the entropy scores only to rank candidate pseudo-labels, not as calibrated probabilities. Since the absolute scale of OT-derived confidence can vary across datasets and iterations as the backbone, projector, and transport plan evolve, a fixed threshold is difficult to choose robustly. A fixed promotion budget $K$ therefore avoids relying on a brittle global threshold, gives direct control over the precision--coverage trade-off, and keeps the expansion schedule stable across rounds. Each selected image receives as its pseudo-label the word with the highest confidence score. These newly labeled items are moved into the aligned set $\mathcal{A}_{k}$ and removed from the unlabeled pool $\mathcal{U}_k$. The training then proceeds to the next round, repeating Phases A--C until all items have been labeled. Importantly, OT is used only to guide which samples to promote during training; inference relies solely on the CTC decoder, making the recognizer lexicon-free at test time.

\section{Results} \label{sec: results}

\textbf{Datasets.}
We evaluate our algorithm on the 
GW
\cite{gw_dataset}, IAM \cite{iam_dataset}, and CVL datasets \cite{cvl_dataset}. Detailed statistics and partitioning conventions for these corpora are available in \cite{wolf2022combining}.

\noindent\textbf{Experimental Setup.}
We pretrain the backbone on 30,000 randomly sampled word images from MJSynth (Synth90k) \cite{MJSynth}, and resize all images to $64\times 256$ pixels. The backbone is optimized with Adam 
at learning rate $10^{-3}$ using the composite loss $\mathcal{L}_{\text{CTC}} + \lambda_{\text{PHOC}}\mathcal{L}_{\text{PHOC}}$, with $\lambda_{\text{PHOC}}=0.5$. In Phase B, the projector is a 3-layer ReLU MLP that maps to a $d=100$ word-embedding space. It is trained with Adam at learning rate $10^{-4}$ using $\mathcal{L}_{\text{proj}} = \mathcal{L}_{\text{sup}} + \lambda_{\text{OT}}\mathcal{L}_{\text{OT}}$, with $\lambda_{\text{OT}}=10^{-2}$ and $\epsilon=0.1$. During Phase A, batches mix synthetic and real samples with dataset-specific ratios $\rho_{\text{syn}}$ (GW 0.5, CVL 0.2, IAM 0). These specific values of $\rho_{\text{syn}}$ reflect dataset scale and the need for regularization. In Phase C, each round promotes a fixed number of high-confidence pseudo-labels per dataset: $K=200$ for GW and $K=5000$ for both CVL and IAM. 

To ensure comparability with prior work, 
we follow the transductive setup of \cite{wolf2022combining},
in the sense of using unlabelled test images in the self-training loop. Concretely, at each iteration $k$ our working pool satisfies $\mathcal{A}_k\cup\mathcal{U}_k = \text{entire dataset (train/val/test)}$.\linebreak We evaluate our algorithm on the official test splits.

\begin{table}[t]
    \centering
    \caption{GW ablation on the PHOC auxiliary head under varying labeled fractions. Lower is better. For the 1\% case, results are averaged over three independent runs to counter variance.}
    \label{tab:gw_phoc_ablation}
    
    \setlength{\tabcolsep}{6pt}
    \begin{tabular*}{\columnwidth}{@{\extracolsep{\fill}}lcccc@{}}
    \hline
    & \multicolumn{4}{c}{\shortstack{Labeled fraction of\\ GW training set}} \\
    \cline{2-5}
    Method / Metric & 1\% & 5\% & 10\% & 100\% \\
    \hline
    \multicolumn{5}{@{}l}{\textbf{CER} $\downarrow$} \\
    CTC only (w/o PHOC) & 31.8 & 6.1 & \textbf{3.0} & 2.6 \\
    CTC + PHOC           & \textbf{11.6} & \textbf{4.8} & 3.2 & \textbf{2.3} \\
    \hline
    \multicolumn{5}{@{}l}{\textbf{WER} $\downarrow$} \\
    CTC only (w/o PHOC) & 44.3 & 11.7 & 6.9 & 7.0 \\
    CTC + PHOC           & \textbf{15.7} & \textbf{8.6} & \textbf{6.7} & \textbf{5.4} \\
    \hline
    \end{tabular*}
\end{table}

\noindent\textbf{Ablation Studies.}
All ablation studies in this paper are conducted on GW only.
To isolate the contribution of the PHOC auxiliary head (see Section \ref{sec: proposed method}), we compare two training configurations for the backbone wherever Eq. \ref{eq: backbone_loss} is applied (see Table \ref{tab:gw_phoc_ablation}).
Concretely, we compare the full objective $\mathcal{L}_{\text{CTC}} + \mathcal{L}_{\text{PHOC}}$ to an ablated variant with $\lambda_\text{PHOC} = 0$ (CTC only), holding architecture, optimization and the downstream pipeline fixed. We report CER and WER on the official GW test split under labeled budgets of $1\%$, $5\%$, $10\%$ and $100\%$.

\begin{figure}[t]
    \centering
    \includegraphics[width=1\linewidth]{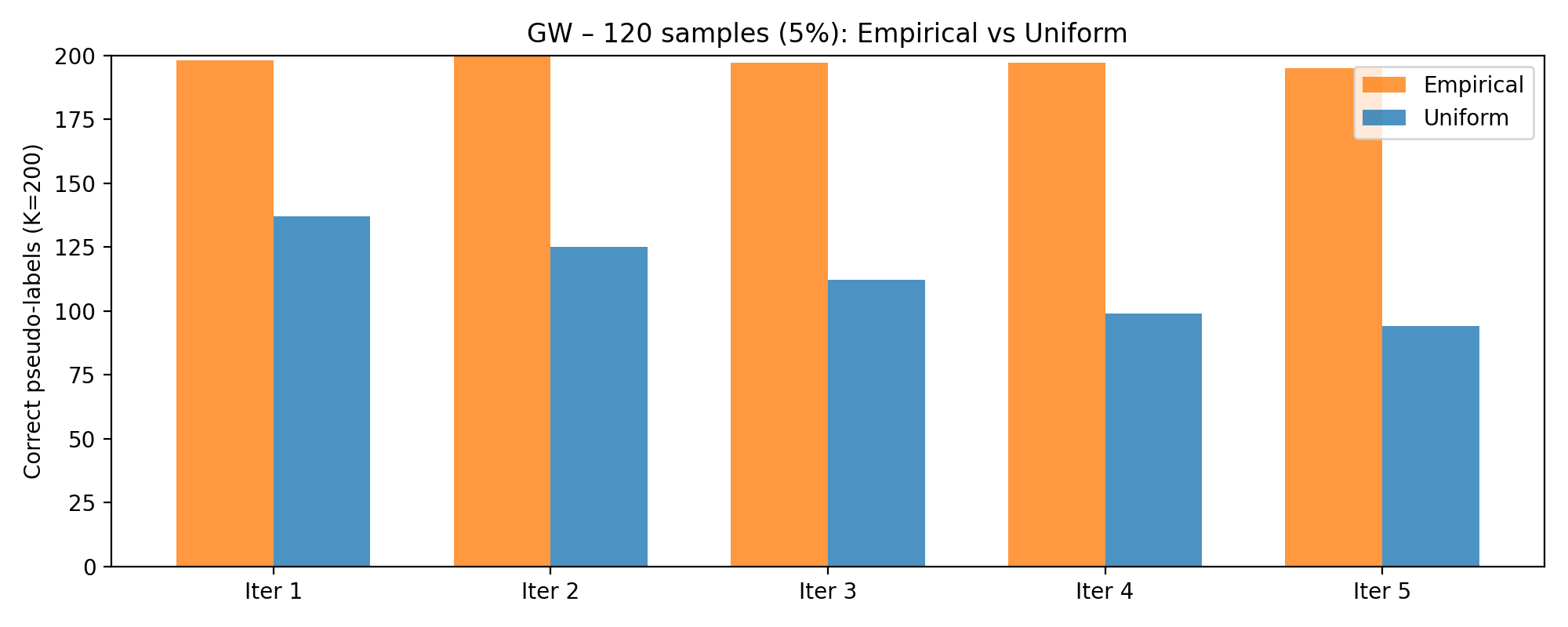}
    \vspace{-25pt}
    \caption{Ablation on the lexical prior $p$ in Phase B OT alignment (GW, 120 labeled $\approx 5\%$; $K=200$ promotions/round). Bars show correct pseudo-labels (of 200) across iterations 1-5 for empirical unigram vs uniform prior.} 
    \label{fig: ablation_lexicon_probs}
\end{figure}

\emph{Ablation on the lexical prior $p$:} 
We ablate the lexicon prior in Phase-B by replacing the empirical unigram over $\mathcal{L}$ with a uniform prior, holding $\lambda_{\text{OT}}$, $\epsilon$, the architecture, and all schedules fixed (see Eq.~\ref{eq: feasible_ot_set}). On GW (120 labels, $\approx\!5\%$) with $K\!=\!200$ promotions per round. Fig.~\ref{fig: ablation_lexicon_probs} reports the number of correct pseudo-labels (of 200) across iterations $k\!=\!1,\dots,5$. The empirical prior keeps promotion precision near ceiling, whereas the uniform prior lowers precision and deteriorates over time. Because balanced OT constrains column masses to $p$, a uniform prior disperses mass toward rare entries, weakening couplings around frequent lexical modes and compounding errors; the Zipf-skewed empirical prior concentrates transport where the data lie, yielding more accurate and stable pseudo-labels.

\begin{figure}[t]
    \centering
    \includegraphics[width=1\linewidth]{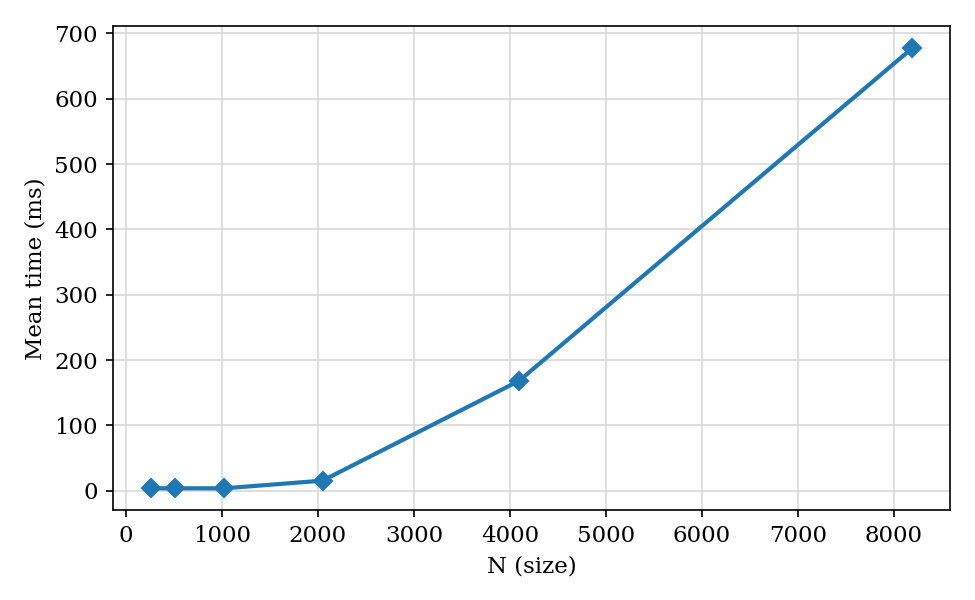}
    \caption{Mean runtime (ms) for computing the optimal transport (OT) loss as a function of a square matrix $N$. Computations are performed on a GPU. Each point reports the average over 300 independent runs.}
    \label{fig:benchmarking_ot}
\end{figure}

\emph{OT runtime:} Figure~\ref{fig:benchmarking_ot} reports the mean runtime of OT loss computation as a function of the square cost-matrix size $N\times N$ (300 runs per $N$), measured on an NVIDIA GeForce RTX 4060 Ti GPU. For the smaller datasets (GW and CVL), the corresponding $N$ falls in a regime where OT can be computed over the full dataset in a single pass; thus, we use full-set OT during training. For IAM, we compute the OT loss in a batched manner by splitting the image embeddings into batches and aggregating the batch-wise OT losses across the dataset. Batched OT should therefore be understood as a computational approximation to the full-dataset coupling. Unlike full-set OT, each mini-batch enforces the lexical marginal only locally and therefore does not exactly preserve the global frequency constraint over the entire dataset. This approximation is used only for IAM, where the vocabulary and dataset size make full-set OT more costly.

\noindent\textbf{Comparisons.}
Table \ref{tab:gw_cvl_results} compares our approach with prior work on GW, CVL and IAM, across labeled data budgets. On GW and CVL, our full system (Ours (T)) achieves the lowest CER/WER across all settings, with the largest gains in the low-label regimes. On IAM, performance is competitive but not state of the art. We conjecture this gap stems from IAM’s much larger vocabulary and higher type–token ratio, which reduce repeated word occurrences, and from the fact that our pretraining uses only 30k samples from the MJSynth synthetic corpus—a very small subset that provides limited coverage of IAM’s lexical 
and writer diversity (657 writers).

\begin{table}[h]
\centering
\caption{Results on GW, IAM, and CVL under varying labeled data ratios. Lower is better.}
\label{tab:gw_cvl_results}

\begingroup
\setlength{\tabcolsep}{1.6pt} 
\renewcommand{\arraystretch}{1.03}
\scriptsize 

\begin{tabular*}{\columnwidth}{@{\extracolsep{\fill}}l c ccc ccc ccc@{}}
\toprule
 & & \multicolumn{3}{c}{GW} & \multicolumn{3}{c}{IAM} & \multicolumn{3}{c}{CVL} \\
\cmidrule(lr){3-5}\cmidrule(lr){6-8}\cmidrule(lr){9-11}
Method &  & 5\% & 10\% & 100\% & 5\% & 10\% & 100\% & 5\% & 10\% & 100\% \\
\midrule
Wolf et al. \cite{wolf2022combining} & \multirow[c]{4}{*}{CER} & 7.5 & 5.9 & 4.0 & \textbf{9.2} & \textbf{8.1} & {6.3} & 5.2 & 4.9 & 2.6 \\
Kang et al. \cite{kang2020unsupervised} &                         & \textemdash & \textemdash & 4.6 & \textemdash & \textemdash & 14.1 & \textemdash & \textemdash & 3.6 \\
Retsinas et al. \cite{retsinas2021iterative} &                     & \textemdash & \textemdash & \textemdash & \textemdash & 9.0 & \textbf{4.0} & \textemdash & \textemdash & \textemdash \\
Ours                         &                         & \textbf{4.8} & \textbf{3.2} & \textbf{2.3} & 13.5 & 11.1 & 6.7 & \textbf{2.8} & \textbf{2.8} & \textbf{1.8} \\
\midrule
Wolf et al. \cite{wolf2022combining} & \multirow[c]{4}{*}{WER} & 20.5 & 17.7 & 12.8 & \textbf{24.9}& \textbf{21.6} & 16.5 & 13.2 & 11.9 & 5.4 \\
Kang et al. \cite{kang2020unsupervised}   &                    & \textemdash & \textemdash & 13.5 & \textemdash & \textemdash & 17.5 & \textemdash & \textemdash & 7.8 \\
Retsinas et al. \cite{retsinas2021iterative} &                 & \textemdash & \textemdash & \textemdash & \textemdash & 28.7 & \textbf{13.9} & \textemdash & \textemdash & \textemdash \\
Ours                         &                         & \textbf{8.6} & \textbf{6.7} & \textbf{5.4} & 27.0 & 24.0 & 15.7 & \textbf{6.2} & \textbf{6.0} & \textbf{3.7} \\
\bottomrule
\end{tabular*}
\endgroup
\end{table}


\section{Conclusion and Future Work} 
\label{sec:conclusion}

We presented a novel framework that introduces key novelties in the traditional HTR paradigm.
Our contribution translates into a method that can lead to a very tangible boost in performance, especially when facing a low-resource regime
of scarcely annotated datasets.
Unlike previous works, our model can leverage mild prior knowledge of lexical structure of the target dataset.
We show that this is possible by reframing the traditional HTR framework
into that of alignment between a visual subspace and a semantic subspace,
carried out efficiently through Optimal Transport.
Compared to other state-of-the-art methods, we have achieved numerical accuracy that ranges from being in the ballpark of other methods (in the larger IAM benchmark) to improving error rate benchmarks by more than $10\%$.
This is the case even using components without ``bells and whistles''
(for example, our pretraining mechanism uses only a very limited training set of $30k$ samples, compared to pretraining with \emph{millions} of tokens in compared work \cite{wolf2022combining}).
We leave further optimization of model components, such as the pretrained model and
loss functions, as future work, along with a controlled comparison to recent 
HTR architectures such as \cite{coquenet2023dan} under matched low-label, pretraining, unlabeled-data, and transductive protocols.

\section{Acknowledgments}
This project is funded by the European Union under Horizon Europe (grant No. 101136568 - HERON).
\begin{center}
    \includegraphics[width=0.35\linewidth]{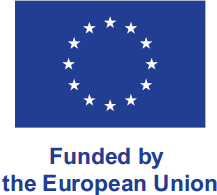}
\end{center}

\newpage

\bibliographystyle{IEEEtran}
\bibliography{refs}

\FloatBarrier
\end{document}